\documentclass[11pt,a4paper]{article}

\usepackage[T1]{fontenc}
\usepackage[utf8]{inputenc}
\usepackage[margin=27mm]{geometry}
\usepackage{mathpazo}
\usepackage{microtype}
\usepackage{url}
\usepackage{tabularx}
\usepackage{array}
\usepackage{graphicx}
\usepackage[authoryear,round]{natbib}
\usepackage[hidelinks,unicode]{hyperref}

\let\cite\citep
\hypersetup{
  pdftitle={Can We Triage LLM Translation Errors in Classical Texts Without Human References? Source Novelty, GEMBA Scoring, and Budgeted Review through Pali-to-English Translation},
  pdfauthor={M\'at\'e Metzger},
  pdfsubject={Reference-free translation quality estimation},
  pdfkeywords={LLM translation, translation quality estimation, classical-language NLP, GEMBA scoring, classical languages}
}

\title{Can We Triage LLM Translation Errors in Classical Texts Without Human References?\\[0.5em]
\large Source Novelty, GEMBA Scoring, and Budgeted Review through Pali-to-English Translation}
\author{M\'at\'e Metzger\\[0.3em]\normalsize Independent Researcher, Hungary}
\date{}

\begin{document}
\maketitle

\begin{abstract}
As large language models become capable translators of classical texts, the binding constraint shifts from producing fluent output to deciding which outputs need expert review, especially where no human reference translation exists. Whether such translations can be triaged for error risk using no reference at inference time is tested through Pali-to-English translation, a demanding testbed: a canonical classical language, low-resource for modern NLP yet backed by a large, segment-aligned source-translation corpus suitable for calibration. An existing human translation serves only to calibrate and validate the detector, never to compute its signals. Three LLMs translated 15,493 passages, and five reference-free signals were compared: source novelty, source-candidate embedding distance, peer-translation disagreement, English-to-Pali backtranslation, and no-reference GEMBA scoring. These signals were calibrated on a 3,000-item reference-informed LLM-adjudicated sample and checked against a 500-item author-adjudicated anchor. Source novelty proved a useful source-side risk prior but not a per-candidate error detector: rare, less formulaic passages failed more often, yet novelty alone missed many candidate-specific errors. Peer-translation disagreement and backtranslation added only secondary signal. The strongest method was no-reference GEMBA scoring by a separate panel of models generally regarded as stronger than the translators: reviewing the top 10\% by GEMBA risk caught 81.6\% of panel-major errors in the LLM-adjudicated calibration set, and these scores remained the best reference-free signal when validated against a human-adjudicated anchor. A same-tier panel, judges from the translators' own class, none scoring its own output, stayed useful but clearly worse, indicating the signal depends on the judge being stronger than the audited system, not on the GEMBA prompt alone. A budgeted triage workflow is proposed: prioritize by source novelty, exploit peer disagreement among candidate translations, and reserve a stronger no-reference judge for candidate-aware review. The workflow is meant to transfer to other classical-to-modern settings, such as Latin, Ancient Greek, and Sanskrit, where references exist for calibration but not for newly translated texts. Confirming that transfer is left to future work.
\end{abstract}

\noindent\textbf{Keywords:} LLM translation; translation quality estimation; classical-language NLP; GEMBA scoring; classical languages

\section{Introduction}

Large language models (LLMs) now produce fluent translations for texts far outside the high-resource modern-language pairs on which machine translation research has historically concentrated. This creates a new practical problem for classical-language scholarship. Many classical languages preserve vast corpora, but they are low-resource languages for modern NLP, and the number of people able to translate them with philological competence is small. As a result, large amounts of classical material remain untranslated or only partially translated. Much of this material is also philosophically or religiously dense: translation often requires domain expertise, while still allowing multiple defensible interpretations, registers, and English renderings. A model translation may read well, and may even agree with a plausible English rendering in broad outline, while still omitting a doctrinally important phrase, reversing agency, mistranslating a technical term, or silently smoothing away a difficult construction. The question is not only whether LLMs can translate classical texts. It is how scholars can triage their outputs when the most valuable use case is precisely the one in which expert translators are scarce and no human reference translation exists for the passage being newly translated.

This paper studies that problem through Pali-to-English translation. Pali is a Middle Indo-Aryan language and the classical language of the Theravada Buddhist canon, or Tipitaka. The Pali Canon preserves one of the most extensive bodies of early Buddhist literature and remains central to Buddhist studies, monastic education, meditation practices, and access to Buddhist thought \cite{vonhinuber1996handbook,gethin1998foundations,gombrich2006theravada,bodhi2005buddha}. It is also a challenging translation domain. Many passages are highly formulaic, while others are elliptical, syntactically dense, or dependent on technical terms whose English renderings are interpretive rather than mechanical.

Pali is therefore a useful stress case for reference-free translation error triage. It is lower-resource than Latin, Ancient Greek, or Sanskrit in the sense relevant to current natural language processing: there are fewer dedicated computational models, benchmarks, and widely used NLP tools, despite the religious and scholarly importance of the corpus. At the same time, Pali has an unusually useful digital infrastructure for this kind of experiment. SuttaCentral and its Bilara data model provide segmented JSON source texts and corresponding English translations at large scale \cite{suttacentral2026about,suttacentralbilara2026}. This combination is rare: a classical language with substantial cultural importance, limited dedicated NLP tooling, and a large segment-aligned source-translation corpus suitable for calibration.

The central task in this study is no-reference at inference time, not no-reference in the entire research pipeline. We use Bhikkhu Sujato's English translation as an adjudication aid when calibrating and validating the detector. We do not use that reference to compute the risk signals; the detector uses only information that would be available for a newly translated passage without a human reference. The proposed inference-time input is only the Pali source and one or more AI candidate translations. The research-time references are used to ask whether the reference-free signals actually correlate with translation errors.

This framing matters because conventional automatic translation evaluation is poorly matched to the humanistic problem. Reference-based metrics are useful when the goal is to compare systems against a known target, but classical translation is not exhausted by agreement with one reference. Translation studies has long emphasized that translation is interpretive, situated, and shaped by audience, purpose, and tradition \cite{berman1992experience,venuti2008invisibility,tymoczko2007enlarging,bassnett2013translation,munday2016introducing}. In Buddhist translation, this is not an abstract point: choices around doctrinal terms such as \texttt{dukkha}, \texttt{dhamma}, \texttt{sa\.{n}kh\={a}ra}, or \texttt{nibb\={a}na} affect how readers understand doctrine. A good audit method must therefore distinguish error from legitimate variation.

The contribution of this paper is an empirical test of reference-free triage signals for Pali-to-English LLM translation. It builds on multi-reference Pali benchmarking work that uses existing human translations to evaluate model outputs \cite{metzger2026palibench}, but asks a different operational question: what can be done when such references are unavailable at inference time? We translate 15,493 Pali passages with three LLMs, compute source-side and candidate-side risk metrics, calibrate those metrics against a 3,000-item reference-informed LLM-adjudicated sample, and validate the main conclusions against a 500-item author-adjudicated anchor set. The goal is not automatic certification of correctness. It is selective review: to estimate which passages most deserve human attention when expert review time is limited.

\section{Literature Review}

\subsection{Translation Variation, Classical Texts, and Buddhist Philology}

The translation of classical and religious texts is not a simple transfer of stable propositions into another language. Translation theory has repeatedly challenged the idea that a translation can be evaluated only as proximity to a single target string. Berman foregrounds the ethical and interpretive difficulty of receiving the foreign text without domestication \cite{berman1992experience}. Venuti argues that fluent translation can conceal the translator's interpretive labor \cite{venuti2008invisibility}, while Tymoczko and Bassnett emphasize that translation choices are embedded in cultural, political, and institutional settings \cite{tymoczko2007enlarging,bassnett2013translation}. Munday's survey of translation studies similarly treats equivalence as only one problem among many, rather than a neutral criterion \cite{munday2016introducing}.

These concerns are especially relevant to Pali Buddhist translation. The canon has been transmitted through religious communities and scholarly institutions over centuries, and its modern English translations differ in register, audience, and interpretive style \cite{vonhinuber1996handbook,gethin1998foundations,bodhi2005buddha,gombrich2006theravada}. A plain modern rendering, a doctrinally conservative rendering, and a philologically literal rendering may all be defensible while differing substantially on the surface. This is one reason that multiple references can improve machine translation training in domains where meaningful variation is expected \cite{wu2024multiple}. For the present study, the lesson is methodological: the target should not be matching any single human translation, but a translation supported by the source text.

\subsection{Machine Translation Evaluation and Quality Estimation}

Machine translation evaluation began from reference overlap metrics such as BLEU, which enabled large-scale system comparison but also made target-string similarity a proxy for quality \cite{papineni2002bleu}. Subsequent metrics attempted to soften this dependence through synonym matching, edit distance, and character-level matching, including METEOR, Translation Edit Rate, and chrF \cite{banerjee2005meteor,snover2006ter,popovic2015chrf}. Neural and learned metrics later moved beyond surface overlap by using contextual embeddings or learned human-judgment prediction, including BERTScore, BLEURT, and COMET \cite{zhang2020bertscore,sellam2020bleurt,rei2020comet}. Work on metric reliability has also shown that automatic scores can mislead when used outside the conditions under which they correlate with human judgments \cite{mathur2020tangled,marie2021credibility}.

Quality estimation addresses a different but closely related task: estimating translation quality without a human reference. This is the tradition most directly connected to the present study. OpenKiwi, TransQuest, COMETKiwi, and xCOMET all represent attempts to predict quality or identify errors from source and candidate output, sometimes with fine-grained error information \cite{kepler2019openkiwi,ranasinghe2020transquest,rei2022cometkiwi,guerreiro2024xcomet}. Human evaluation practice has also shifted toward explicit error annotation, especially through Multidimensional Quality Metrics (MQM), because aggregate fluency scores alone do not reveal whether a translation contains consequential meaning errors \cite{lommel2014mqm,freitag2021experts,lommel2024multirange}.

The present study shares the quality-estimation goal but differs in domain and deployment assumptions. Most quality-estimation work is trained and evaluated on modern language pairs with established benchmark data. Pali-to-English translation lacks that infrastructure. Recent domain-specific low-resource QE work shows that closed LLMs can perform strongly with prompting, while more robust open or local QE systems may require adaptation with labeled data \cite{gurav2026quality}. The method tested here therefore uses transparent corpus-derived features, peer-translation disagreement, backtranslation, and LLM-based scoring as reference-free signals that can be calibrated with a sampled reference-aided adjudication set.

\subsection{LLMs as Translation Judges}

Recent work has shown that LLMs can serve as scalable evaluators of generated text, but also that their judgments are not neutral instruments. MT-Bench and related work made LLM-as-a-judge evaluation prominent \cite{zheng2023judging}. G-Eval and Prometheus further showed that LLM evaluators can follow rubrics and produce scores aligned with human judgments in some settings \cite{liu2023geval,kim2023prometheus}. At the same time, LLM judges can show position bias, self-preference, model-family bias, and instability across prompts or rubrics \cite{wang2024fair}. For translation, Kocmi and Federmann introduced GPT Estimation Metric Based Assessment (GEMBA), showing that prompted LLMs can be strong translation-quality evaluators, including in no-reference modes \cite{kocmi2023gemba}. GEMBA-MQM extended this idea toward error-span and MQM-style evaluation without human references \cite{kocmi2023gembamqm}.

This literature motivates the present design but also constrains its claims. We use LLM judges because they make a 3,000-item calibration feasible, but we do not treat them as a gold standard. Their labels are checked against a 500-item author anchor, and the paper separately evaluates whether GEMBA merely predicts judgments made by models or remains useful when compared to author labels.

\subsection{Classical-Language NLP and the Pali Test Case}

Classical languages occupy an awkward place in NLP. They may have large surviving corpora and high cultural importance, yet they are often low-resource for modern computational purposes: fewer annotated datasets, fewer domain-specific models, uneven tokenization, and a shortage of expert-labeled evaluation data. Broader work on linguistic diversity in NLP has shown how sharply language resources are concentrated in a small number of high-resource languages \cite{joshi2020state}, and participatory low-resource MT work emphasizes that data creation and evaluation must be tied to communities and domain experts rather than treated as purely technical extraction \cite{nekoto2020participatory}.

Recent classical-language NLP work shows both the promise and the limits of LLM-assisted philology. Latin BERT demonstrates the value of language-specific contextual models for classical philological tasks \cite{bamman2020latinbert}. LITERA frames Latin-to-English LLM translation as a research-assistance workflow rather than a replacement for expert translators \cite{rosu2025litera}. Work on Ancient Greek medical and philosophical prose shows that LLMs can produce useful translations but fail sharply on rare technical language \cite{zainaldin2026galen}. Sanskrit-English work such as Itihasa demonstrates that even large parallel corpora for premodern Indic texts can remain difficult for standard translation architectures \cite{aralikatte2021itihasa}.

Pali has received less dedicated computational attention than Latin, Ancient Greek, or Sanskrit, though digital Buddhist studies projects have made important infrastructure available. Zigmond's computational analysis of the Pali Canon demonstrates the feasibility of corpus-level quantitative work on Pali texts \cite{zigmond2021computational}. Prior work on AI-driven translation of ancient Buddhist scriptures used GEMBA and other metrics to evaluate Pali-to-English LLM translations in a narrower scripture-translation setting \cite{phophichit2026quality}. PaliBench shows how independently translated Pali passages can be turned into a multi-reference benchmark for classical-language translation \cite{metzger2026palibench}. SuttaCentral provides early Buddhist texts, translations, parallels, and the Bilara translation infrastructure, including segment-level JSON data used in this study \cite{suttacentral2026intro,suttacentralbilara2026}. This makes Pali a particularly informative test case: under-resourced enough to expose the difficulty of classical-language translation, but structured enough to support a large empirical audit.

The gap addressed here is therefore narrow but important. Existing MT evaluation research gives tools for reference-based scoring, quality estimation, error annotation, and LLM judging. Existing classical-language NLP shows that LLM translation can be useful but fragile. What is missing is a practical, empirically calibrated way to triage AI translations of classical texts when no human reference translation is available at the point of use. This paper tests that missing layer.

\section{Methodology}

\subsection{Study Design}

This study evaluates whether Pali-to-English large language model (LLM) translations can be triaged for error risk without using human reference translations at inference time. The practical target is not automatic certification of correctness. The target is selective review: given a large set of AI translations, can a reference-free system rank or flag those most likely to contain correction-worthy errors, so that a limited human review budget is spent where it matters most? This places the study within the broader quality-estimation tradition in machine translation, where the goal is to estimate translation quality from source and candidate output rather than from candidate-reference overlap \cite{rei2022cometkiwi,guerreiro2024xcomet}.

The design separates three tasks that are often conflated in automatic translation evaluation. First, candidate translations are produced from Pali source passages. Second, no-reference risk signals are computed from the source passage, the candidate translation, and, for some signals, other AI translations of the same passage. Third, a sampled subset is adjudicated using a reference-aided procedure so that the no-reference signals can be calibrated against error labels. In other words, human reference translations are used to evaluate the detector, not to compute the detector.

The study is deliberately framed around budgeted triage, but not as a simple cheap-versus-expensive binary. The signals tested here differ in what they require. Source novelty is a source-only prior and can be computed before translation. Source-candidate embedding distance requires the candidate translation and embeddings. Peer-centroid distance is no-reference in the human-translation sense, but it is not intrinsically cheap because it requires additional AI translations of the same passage. Backtranslation and GEMBA-style scoring require further LLM calls and are therefore higher-effort candidate-aware checks. This triage framing reflects the intended use case for classical-language translation: many passages may be translated, but only a small proportion can receive intensive expert attention.

Although implemented here for Pali-to-English translation, the methodological unit is more general: source passage, candidate translation, optional peer translations, and a calibration set with adjudicated error labels. The Pali-specific parts of the workflow are the tokenizer, source-corpus rarity estimates, and philological adjudication criteria. The same design can therefore be adapted to other classical-to-modern translation settings where enough source text is available to estimate source difficulty and at least a sampled set can be adjudicated.

\subsection{Corpus Construction}

The main corpus was constructed from Bilara JSON files containing Pali source text and Bhikkhu Sujato's corresponding English translations, available through SuttaCentral's public GitHub repository under permissive licensing. Bilara stores texts at segment level. A segment is one keyed unit such as \texttt{mn1:1.1} or \texttt{mn1:1.2}. A passage is the top-level key obtained by removing the final segment component after the last dot: for example, \texttt{mn1:1.1} and \texttt{mn1:1.2} both belong to passage \texttt{mn1:1}. In this study, the passage, not the segment, is the translation and evaluation unit, because passages are usually more semantically complete and interpretable while remaining short enough for controlled model calls and adjudication.

All Pali segments belonging to a passage were concatenated in Bilara segment order. The corresponding English segments (Bhikkhu Sujato's translations) were concatenated in the same way for reference-aided adjudication. Title and front-matter passages ending in \texttt{:0} were excluded. The final corpus was then filtered to remove very short, incomplete, or anomalously aligned units before translation. The filtering rules are shown in Table 1.

\begin{table}[tbp]
\caption{Final corpus filtering rules.}
\begin{center}
\small
\renewcommand{\arraystretch}{1.15}
\begin{tabularx}{\textwidth}{@{}>{\raggedright\arraybackslash}X >{\raggedright\arraybackslash}X@{}}
\hline
\textit{Rule} & \textit{Reason} \\
\hline
Exclude passage ids ending in \texttt{:0} & Removes title and front-matter units rather than substantive translated passages. \\
Require non-empty Pali and English passage text & Removes unusable source or reference records. \\
Require at least 10 normalized Pali content-word tokens & Removes very short passages that provide little information for translation-error evaluation. \\
Require every Pali segment in the passage to have a corresponding English segment & Avoids incomplete source-reference alignment. \\
Require English/Pali character ratio between 0.5 and 2.0 & Removes likely extraction, segmentation, or alignment anomalies. \\
\hline
\end{tabularx}
\end{center}
\end{table}

The source profile contained 26,165 eligible non-title passages before final filtering. The final test corpus contains 15,493 passages and 91,474 source segments, retaining 59.2\% of the profiled non-title passage pool. Non-exclusive rejection counts were 6,416 passages under the 10-content-token threshold, 3,755 passages with incomplete English segment coverage, and 1,775 passages outside the English/Pali character-ratio bounds. On the Pali source side, the final corpus contains 6.48 million characters, 725,623 whitespace-delimited words, and 3.73 million \path|cl100k_base| tokens, roughly equivalent to 1,450-1,600 printed A4 pages depending on words-per-page assumptions; Sujato's corresponding English passages were retained as reference aids for adjudication.

\subsection{Source Novelty and Corpus Stratification}

A central hypothesis of the study is that error risk is partly conditioned by source difficulty. For this reason, every Pali passage was assigned a continuous source novelty index before sampling. The index is source-only: it can be computed before any AI translation is generated. This choice is also motivated by recent ancient-language translation work in which terminology rarity was a strong predictor of LLM translation failure \cite{zainaldin2026galen}.

Normalized content tokens were produced by Unicode normalization, lowercasing, normalizing \texttt{\.{m}} to \texttt{\d{m}}, extracting word-like Pali tokens, dropping tokens shorter than four characters, and excluding a small set of common particles such as \texttt{atha}, \texttt{ca}, \texttt{eva}, \texttt{eva\d{m}}, \texttt{iti}, \texttt{kho}, \texttt{pana}, \texttt{pi}, and \texttt{v\={a}}. Source novelty combines four z-scored source-side components:

\begin{enumerate}
\item content-word rarity, measured as the density of normalized content tokens with document frequency $\leq$ 10 in the Pali corpus;
\item character 5-gram rarity, measured as the density of normalized character 5-grams with document frequency $\leq$ 3;
\item formulaicity, measured by selected repeated Pali 5-10 word n-grams;
\item source-neighborhood density, measured as the top-5 mean similarity to other Pali passages using character 3-5 gram TF-IDF cosine similarity.
\end{enumerate}

The index is computed as:

\begin{verbatim}
source_novelty_index =
    (
      z(content_word_df10_rarity)
    + z(char5gram_df3_rarity)
    - z(repeated_formula_count)
    - z(top5_source_neighborhood_density)
    ) / 4
\end{verbatim}

Higher values indicate passages that are rarer, less formulaic, and more isolated from near-neighbor source passages. The final corpus was divided into low, middle, and high novelty bands for descriptive analysis and stratified sampling. These bands are not treated as natural categories; the continuous novelty score is the primary source-side measure. Corpus counts by novelty band are reported in Appendix Table A1.

\subsection{Candidate Translation Generation}

Three OpenRouter-hosted translator models were selected for the main experiment: \path|deepseek/deepseek-v4-pro|, \path|qwen/qwen3.6-plus|, and \path|x-ai/grok-4.3|. The aim was to use capable models that could plausibly produce useful translations, but not the strongest and most expensive frontier systems. This was a deliberate calibration choice: the study requires enough errors to evaluate triage behavior, while still keeping full-corpus translation practically affordable. Prior PaliBench results also informed the choice of model families by identifying systems that were relatively strong but still produced measurable high-drift outliers relative to human-reference consensus, some of which are expected to correspond to real errors \cite{metzger2026palibench}. The design is therefore about evaluating error-risk triage, not constructing a leaderboard of the best possible Pali translators.

Each model translated all 15,493 Pali passages using the same passage-level translation prompt. The prompt instructed the model to translate into English only, preserve named persons, places, numbers, lists, negation, agents, and doctrinal terms, and return JSON output. The exact translator, adjudication, backtranslation, and GEMBA prompts are included in the reproducibility package as part of the runnable scripts. Passages were batched under an approximate 3,000-token input budget per API call. The translation script supported interruption and resumption. It also detected source-copy-like failures, where the model output substantially reproduced the Pali source instead of translating it. These occurred frequently during Grok generation; the script retried them with smaller batches and retained persistent failures as candidate outputs rather than silently deleting them.

DeepSeek and Qwen produced complete translation files for all 15,493 passages. Grok produced 15,491 ordinary translations and two persistent source-copy failures. These two failures were retained because copying the Pali source instead of translating it is a real translation failure and should be visible to a triage system. The full translation table therefore contains 46,479 translation instances, one for each translator model - passage pair.

\begin{table}[tbp]
\caption{Translator models and candidate output counts.}
\begin{center}
\small
\renewcommand{\arraystretch}{1.15}
\begin{tabularx}{\textwidth}{@{}>{\raggedright\arraybackslash}X >{\raggedright\arraybackslash}X >{\raggedright\arraybackslash}X@{}}
\hline
\textit{Translator model} & \textit{Output instances} & \textit{Notes} \\
\hline
\path|deepseek/deepseek-v4-pro| & 15,493 & Complete translation file. \\
\path|qwen/qwen3.6-plus| & 15,493 & Complete translation file. \\
\path|x-ai/grok-4.3| & 15,493 & Includes two persistent source-copy failure candidates. \\
\textbf{Total} & \textbf{46,479} & Three candidate translations for each Pali passage. \\
\hline
\end{tabularx}
\end{center}
\end{table}

\subsection{Embeddings and Full-Corpus Feature Table}

All Pali source passages, English references, and candidate translations were embedded using \path|google/gemini-embedding-2-preview| through OpenRouter. This model family was selected because Gemini embeddings report strong performance on large embedding benchmarks such as MTEB/MMTEB, including multilingual and retrieval-oriented tasks, and because preliminary retrieval tests indicated better behavior than the other candidate embedding models tested \cite{muennighoff2023mteb,lee2025gemini}. The resulting SQLite cache contains 77,465 embeddings: Pali source, English reference, DeepSeek candidate, Qwen candidate, and Grok candidate for each of the 15,493 passages.

The full-corpus feature table contains one row per translation instance. The features fall into five groups:

\begin{enumerate}
\item \textbf{Source-only features:} source novelty index and related rarity/formulaicity components.
\item \textbf{Candidate-source features:} embedding distance between Pali source and English candidate, and source/candidate length ratios.
\item \textbf{Peer-consensus features:} distance from the candidate translation to the centroid of the other two AI translations of the same passage.
\item \textbf{Reference-aided diagnostic features:} distance between candidate translation and English (Sujato) reference, used only for analysis and not as a no-reference detector.
\item \textbf{Hard anomaly flags:} persistent source-copy or structurally invalid translation failures.
\end{enumerate}

Peer-centroid distance is a no-human-reference signal. For each passage and candidate model, the other two AI translations are embedded and averaged to form a peer centroid. The candidate's cosine distance from that centroid measures how much it diverges from the other AI renderings. This use of peer translations is related to multi-hypothesis MT evaluation, where model-output variability can substitute for or complement human-reference variation \cite{fomicheva2020multihypothesis}. It is a candidate-specific adequacy proxy, but it is not a low-effort signal in a single-translation deployment because it requires multiple translations of the same source passage. A normalized peer-drift variant was also computed, but the raw candidate-to-peer-centroid distance is preferred because only two peer translations are available and the denominator of a normalized two-peer envelope can be unstable.

\subsection{Calibration Sample}

The 46,479 translation instances were ranked by an initial risk score:

\begin{verbatim}
z(source_novelty_index)
+ z(candidate_distance_to_peer_centroid)
+ z(source_candidate_distance)
\end{verbatim}

Hard anomaly flags (e.g. source-copy failures) were forced into the top-ranked tail. This preliminary score was used only to construct a calibration sample broad enough to estimate both high-risk and low-risk behavior. It was not treated as the final detector.

The calibration set contains 3,000 translation instances and was sampled with fixed seed \texttt{20260519}. The top 500 highest-risk instances were included deterministically as a very-high-risk stratum. The remaining 2,500 items were sampled as 500-item strata from the rest of the top 20\%, the 20-40\% range, the 40-60\% range, the 60-80\% range, and the bottom 20\% of the full risk ranking. Sampling was stratified by translator model and novelty band where possible. This design deliberately overrepresents high-risk material while still covering the full score distribution.

\begin{table}[tbp]
\caption{Calibration sample design.}
\begin{center}
\small
\renewcommand{\arraystretch}{1.15}
\begin{tabularx}{\textwidth}{@{}>{\raggedright\arraybackslash}X >{\raggedright\arraybackslash}X >{\raggedright\arraybackslash}X >{\raggedright\arraybackslash}X@{}}
\hline
\textit{Calibration stratum} & \textit{Definition} & \textit{Pool size} & \textit{Sample size} \\
\hline
Very high & Literal top 500 by initial risk rank, with hard anomalies sorted first & 500 & 500 \\
High & Remaining ranks after top 500 through top 20\% & 8,796 & 500 \\
Upper mid & 20-40\% risk percentile range & 9,296 & 500 \\
Medium & 40-60\% risk percentile range & 9,296 & 500 \\
Lower mid & 60-80\% risk percentile range & 9,296 & 500 \\
Low & Bottom 20\% & 9,295 & 500 \\
\textbf{Total} &  & \textbf{46,479} & \textbf{3,000} \\
\hline
\end{tabularx}
\end{center}
\end{table}

The calibration sample contains 987 DeepSeek translations, 1,030 Qwen translations, and 983 Grok translations. By source novelty band, it contains 863 low-novelty, 852 middle-novelty, and 1,285 high-novelty instances. The high-novelty overrepresentation is expected because high novelty was part of the initial risk score and the top-ranked tail was intentionally enriched. Counted item-wise, the 3,000 sampled translation instances contain 17,104 Pali source segments, 1.16 million source characters, 129,902 whitespace-delimited source words, and 670,300 \path|cl100k_base| source tokens, while representing 2,587 unique Pali passages because some passages are sampled with more than one candidate model.

\subsection{Error Adjudication}

Each calibration item was adjudicated by three LLM judges: \path|openai/gpt-5.5|, \path|google/gemini-3.1-pro-preview|, and \path|anthropic/claude-sonnet-4.6|. Judges saw the Pali source, the English reference as an adjudication aid, and the candidate English translation. They did not see the translator model identity, risk score, novelty band, GEMBA score, or any other metrics. This follows the growing use of strong LLMs as scalable evaluators while retaining the need to check them against human judgment because LLM evaluators can exhibit systematic biases \cite{zheng2023judging,wang2024fair}.

The judge task was binary at the primary level: decide whether the candidate is a valid translation variation or contains a translation error. If an error was present, the judge also assigned severity:

\begin{itemize}
\item \textbf{Minor error:} a local or limited issue that would be worth correcting but does not materially mislead the reader about the passage's main meaning.
\item \textbf{Major error:} an error that materially changes the meaning, omits essential content, reverses polarity, assigns agency or roles incorrectly, adds unsupported content, or mishandles an important doctrinal or contextual term.
\end{itemize}

Judges returned structured JSON. One API call was made per judge per passage. The full calibration adjudication therefore used 9,000 judge calls. All calls completed with valid schema output.

Panel labels were aggregated by majority vote. A translation was labeled \texttt{ERROR} if at least two judges labeled it as an error. Among panel-error items, it was labeled \texttt{MAJOR} if at least two judges assigned major severity; otherwise it was labeled \texttt{MINOR}. A translation was labeled \texttt{VALID} if fewer than two judges labeled it as an error.

\subsection{Source-Prior and Embedding-Based Triage Signals}

The first group of triage signals does not invoke additional LLM judges after translations and embeddings are available. Three signals were evaluated singly and in combination: source novelty as a source-only difficulty prior, source-candidate embedding distance as a candidate-aware source-to-translation distance, and peer-centroid distance as a measure of divergence from the other AI translations of the same passage. These signals differ in deployment cost: source novelty can be computed before translation, source-candidate distance requires one candidate translation and embeddings, and peer-centroid distance requires multiple candidate translations.

The initial equal-weight score used all three features. Subsequent exploration tested each feature alone and transparent z-scored linear combinations. A simple cross-validation-selected source-novelty-plus-peer score was retained as the main refined routing rule:

\begin{verbatim}
refined_source_peer_score =
    0.7 * z(source_novelty_index)
  + 0.3 * z(candidate_distance_to_peer_centroid)
\end{verbatim}

The weight search evaluated simple z-scored linear combinations of source novelty, peer-centroid distance, and source-candidate distance, selecting weights by held-out major-error recall at fixed review budgets. Grouped cross-validation by Pali passage id was used so that translations of the same Pali passage could not appear in both training and held-out folds. Small learned models, including logistic regression and shallow tree-based classifiers, were also tested as exploratory baselines but were not chosen as the primary method because they did not clearly outperform the transparent novelty-heavy score. A small exploratory run with a public pruned COMETKiwi no-reference QE checkpoint produced near-random ranking performance on the calibration and author-anchor labels, so it was not retained as a main baseline.

\subsection{Higher-Effort Candidate-Aware Checks}

The second group contains higher-effort reference-free checks intended for cases where a passage has already been routed for deeper scrutiny, or where budget allows a more intensive audit. These checks are candidate-aware and require additional LLM calls.

\subsubsection{Backtranslation}

Each calibration candidate was translated back from English into Pali using \path|openai/gpt-5.5|. The backtranslation prompt instructed the model to translate the meaning of the candidate translation, avoid reconstructing unseen source text, preserve named persons, places, numbers, lists, negation, agents, and doctrinal terms, and return only a JSON object containing the Pali backtranslation. This produced complete backtranslations for all 3,000 calibration items.

Backtranslation risk was measured by comparing the original Pali source to the backtranslated Pali. Several lexical metrics were tested: chrF risk, character 5-gram weighted Jaccard risk, content-token weighted Jaccard risk, and recall-like variants focused on rare tokens or n-grams. Higher risk means the backtranslation preserves less source-specific Pali material.

\subsubsection{GEMBA No-Reference Scoring}

GEMBA-style no-reference scoring asks an LLM to assign a direct quality score to a translation given only the source and candidate translation. The prompt was based on the no-reference GEMBA-DA prompt \cite{kocmi2023gemba}, with an added numeric-only output constraint for reliable parsing. Scores were interpreted on a 0-100 scale where higher scores indicate better translation quality. Risk features were then computed as \texttt{100 - score} or corresponding summaries across multiple scorers.

We used GEMBA-DA rather than GEMBA-MQM because the present task is budgeted triage: translations must be ranked by review priority under fixed review fractions. A scalar direct-assessment score is therefore easier to calibrate as a risk signal than MQM annotation. GEMBA-MQM is closely related prior work for diagnostic error-span detection without references \cite{kocmi2023gembamqm}, but in this study diagnostic labeling is handled separately by the LLM adjudication panel and the author anchor.

Two GEMBA panels were tested. The "strong" GEMBA panel used \path|openai/gpt-5.5|, \path|google/gemini-3.1-pro-preview|, and \path|anthropic/claude-sonnet-4.6|. The "peer" GEMBA panel used the translator models themselves where possible, excluded self-scoring, and added \path|moonshotai/kimi-k2.6| so that each candidate still received three "peer" panel scores. Thus, a DeepSeek translation was scored by Qwen, Grok, and Kimi; a Qwen translation by DeepSeek, Grok, and Kimi; and a Grok translation by DeepSeek, Qwen, and Kimi.

For each panel, the primary GEMBA feature is mean risk across the three available scores. Several derived features were also computed as exploratory checks, including median risk, lower-two mean risk, minimum-score risk, maximum-score risk, and score range. Mean risk is the main reported ranking signal; the other summaries test whether alternative aggregation or scorer disagreement adds information beyond the average score.

\subsection{Author Anchor Set}

Because the 3,000-item calibration labels are produced by LLM judges, a 500-item author-adjudicated anchor set was created with fixed seed \texttt{20260523} to test whether the LLM panel and GEMBA risk signals align with human judgment. The anchor set was diagnostically enriched rather than prevalence-balanced. It was sampled to stress the most important boundary cases: "strong" GEMBA high-risk items, "strong" GEMBA low-risk items, LLM panel errors outside the "strong" GEMBA top 20\%, "strong" versus "peer" GEMBA disagreements, high-source-novelty items with low GEMBA risk, and random stratified controls.

The author adjudicator has formal training in Pali and saw the same core evidence as the LLM judges: Pali source, Sujato reference aid, and candidate translation. The author anchor set was labeled using the same three outcome labels: \texttt{VALID}, \texttt{MINOR}, and \texttt{MAJOR}.

\begin{table}[tbp]
\caption{Author anchor sampling design.}
\begin{center}
\small
\renewcommand{\arraystretch}{1.15}
\begin{tabularx}{\textwidth}{@{}>{\raggedright\arraybackslash}X >{\raggedright\arraybackslash}X@{}}
\hline
\textit{Anchor group} & \textit{Items} \\
\hline
"strong" GEMBA high risk and LLM panel major & 80 \\
"strong" GEMBA high risk and LLM panel valid or minor & 70 \\
Outside "strong" GEMBA top 20\% but LLM panel error & 100 \\
"strong" GEMBA low risk and LLM panel valid & 70 \\
"strong" versus "peer" GEMBA disagreement & 80 \\
High source novelty but low "strong" GEMBA risk & 50 \\
Random stratified controls & 50 \\
\textbf{Total} & \textbf{500} \\
\hline
\end{tabularx}
\end{center}
\end{table}

The anchor set is not used to estimate corpus prevalence. Its purpose is to test whether the LLM panel and the strongest risk metrics remain informative when checked against a human adjudicator.

\subsection{Statistical Analysis}

The main labels are three-class severity labels (\texttt{VALID}, \texttt{MINOR}, \texttt{MAJOR}) and binary reductions (\texttt{ERROR} versus \texttt{VALID}, \texttt{MAJOR} versus non-major). Error-rate estimates are reported as proportions with Wilson score intervals where appropriate. Stratified full-corpus prevalence estimates weight each calibration stratum by its actual size in the 46,479-instance corpus.

Ranking metrics are evaluated by area under the receiver operating characteristic curve (AUC), review-budget recall, and review precision. AUC is used as a threshold-independent ranking measure: it estimates how often a randomly chosen error receives a higher risk score than a randomly chosen non-error. Review-budget curves ask: if only the top 1\%, 5\%, 10\%, or 20\% of items under a given risk score are reviewed, what share of known major errors are captured, and how many reviewed items are actually errors? Each metric is evaluated under its own ranking: for example, a 10\% GEMBA budget means selecting the highest-risk 10\% by GEMBA score, while a 20\% source-novelty budget means selecting the highest-risk 20\% by source novelty.

Because the calibration design oversamples high-risk items, raw percentages in the 3,000-item calibration set should not be interpreted as corpus prevalence. Stratified estimates are used when making full-corpus claims. Conversely, metric comparisons inside the 3,000-item set are treated as calibrated ranking comparisons, not direct prevalence estimates.

\section{Results}

\subsection{Calibration Labels and Error Enrichment}

The LLM panel labeled 207 of 3,000 calibration items as major errors (6.9\%, 95\% Wilson interval 6.0-7.9\%), 337 as minor errors, and 2,456 as valid variations. Total panel-labeled errors were 544 of 3,000 (18.1\%). Because the calibration sample intentionally oversampled the high-risk tail, these raw rates are not corpus prevalence estimates.

Error rates vary strongly by risk stratum. The top 500 very-high-risk items have a panel-major rate of 28.0\% and any-error rate of 49.6\%. All other strata have much lower major-error rates, ranging from 1.4\% to 4.6\%. The result confirms that the initial no-reference risk ranking concentrated major errors in the extreme tail, but also shows that non-tail strata still contain errors.

\begin{table}[tbp]
\caption{LLM-panel error rates by calibration stratum.}
\begin{center}
\small
\renewcommand{\arraystretch}{1.15}
\begin{tabularx}{\textwidth}{@{}>{\raggedright\arraybackslash}X >{\raggedright\arraybackslash}X >{\raggedright\arraybackslash}X >{\raggedright\arraybackslash}X >{\raggedright\arraybackslash}X@{}}
\hline
\textit{Stratum} & \textit{n} & \textit{Major error \% (95\% CI)} & \textit{Minor error \% (95\% CI)} & \textit{Any error \% (95\% CI)} \\
\hline
Very high & 500 & 28.0\% (24.2-32.1) & 21.6\% (18.2-25.4) & 49.6\% (45.2-54.0) \\
High & 500 & 4.6\% (3.1-6.8) & 12.2\% (9.6-15.4) & 16.8\% (13.8-20.3) \\
Upper mid 20-40 & 500 & 3.4\% (2.1-5.4) & 9.0\% (6.8-11.8) & 12.4\% (9.8-15.6) \\
Medium 40-60 & 500 & 1.4\% (0.7-2.9) & 9.2\% (7.0-12.1) & 10.6\% (8.2-13.6) \\
Lower mid 60-80 & 500 & 2.2\% (1.2-3.9) & 7.2\% (5.2-9.8) & 9.4\% (7.1-12.3) \\
Low bottom 20 & 500 & 1.8\% (0.9-3.4) & 8.2\% (6.1-10.9) & 10.0\% (7.7-12.9) \\
\hline
\end{tabularx}
\end{center}
\end{table}

After weighting each stratum by its full-corpus size, the estimated panel-labeled full-corpus major-error prevalence is 2.9\% (approximate 95\% CI 2.3-3.5\%). The estimated minor-error prevalence is 9.3\% (8.2-10.3\%), and the estimated any-error prevalence is 12.2\% (11.0-13.4\%).

\begin{table}[tbp]
\caption{Stratified full-corpus prevalence estimates from LLM-panel labels.}
\begin{center}
\small
\renewcommand{\arraystretch}{1.15}
\begin{tabularx}{\textwidth}{@{}>{\raggedright\arraybackslash}X >{\raggedright\arraybackslash}X >{\raggedright\arraybackslash}X@{}}
\hline
\textit{Label} & \textit{Estimated count in 46,479 instances} & \textit{Estimated prevalence} \\
\hline
Major error & 1,362.6 & 2.9\% (2.3-3.5) \\
Minor error & 4,304.5 & 9.3\% (8.2-10.3) \\
Any error & 5,667.1 & 12.2\% (11.0-13.4) \\
\hline
\end{tabularx}
\end{center}
\end{table}

Broad risk-boundary estimates first evaluate the original preliminary risk rank used to construct the calibration sample. This rank combined three z-scored components with equal weight: source novelty, candidate distance from the peer-translation centroid, and source-candidate embedding distance, with hard anomaly flags forced into the top-ranked tail. Its purpose was to create a calibration sample enriched for likely errors while still covering the full risk distribution; it was not assumed to be the final detector. Under this original rank, reviewing the top 20\% of the full corpus would capture an estimated 40.0\% of panel-major errors and 30.5\% of all panel errors. Reviewing the top 40\% would capture 63.2\% of panel-major errors. These estimates demonstrate useful enrichment, but also show that the preliminary rank is not sufficient as a final detector. Detailed broad-boundary estimates are reported in Appendix Table A2.

\subsection{Source-Prior and Peer-Translation Routing}

The first group of signals asks a simple question: can we rank translations so that a limited review budget sees more errors than random review would? These signals are "reference-free" in the deployment sense: they do not use a human translation of the passage being audited. They differ, however, in cost. Source novelty is source-only and can be computed before translation. Peer-centroid distance requires several independent AI translations of the same Pali passage and therefore measures candidate disagreement, not source difficulty alone. Source-candidate embedding distance and length anomaly are candidate-level checks, but in this experiment they were weaker.

Table 7 reports stratified full-corpus estimates of recall, precision, and the reviewed non-error share at a 20\% review budget. Source novelty alone captured an estimated 48.7\% of panel-major errors. Peer-centroid distance alone captured 40.9\%. The original equal-weight composite (source novelty, peer-centroid distance, and source-candidate embedding distance) captured about the same amount, 39.8\%, because it gave too much weight to weaker components.

The best transparent routing score was a novelty-heavy combination: 0.7 source novelty plus 0.3 peer-centroid distance. This score captured an estimated 60.6\% of panel-major errors, 31.3\% of minor errors, and 38.4\% of all errors at a 20\% review budget. Its major-error precision was only 8.9\%, so it is not a reliable stand-alone error detector. Its value is as a routing rule: source novelty identifies passages that are intrinsically harder or less formulaic, while peer-centroid distance adds evidence that one candidate translation diverges from other model renderings of the same source.

\begin{table}[tbp]
\caption{Leading source-prior and embedding-based scores at a 20\% review budget.}
\begin{center}
\scriptsize
\renewcommand{\arraystretch}{1.15}
\begin{tabularx}{\textwidth}{@{}>{\raggedright\arraybackslash}X >{\raggedright\arraybackslash}X >{\raggedright\arraybackslash}X >{\raggedright\arraybackslash}X >{\raggedright\arraybackslash}X >{\raggedright\arraybackslash}X >{\raggedright\arraybackslash}X@{}}
\hline
\textit{Score} & \textit{Major recall} & \textit{Minor recall} & \textit{Any-error recall} & \textit{Major precision} & \textit{Any-error precision} & \textit{Reviewed non-error share} \\
\hline
Refined source-novelty-plus-peer score: 0.7 novelty + 0.3 peer & 60.6\% & 31.3\% & 38.4\% & 8.9\% & 23.4\% & 76.6\% \\
Source novelty only & 48.7\% & 35.9\% & 39.0\% & 7.1\% & 23.8\% & 76.2\% \\
Peer-centroid distance only & 40.9\% & 28.4\% & 31.4\% & 6.0\% & 19.2\% & 80.8\% \\
Original equal composite & 39.8\% & 27.4\% & 30.4\% & 5.8\% & 18.5\% & 81.5\% \\
Source-candidate embedding distance only & 36.5\% & 20.4\% & 24.3\% & 5.4\% & 14.8\% & 85.2\% \\
Length anomaly only & 25.9\% & 18.4\% & 20.2\% & 3.8\% & 12.3\% & 87.7\% \\
\hline
\end{tabularx}
\end{center}
\end{table}

Cross-validation supported this weighting direction but also showed that the result should not be overinterpreted. When passages were grouped so that translations of the same Pali passage could not appear in both training and held-out folds, the selected weight was usually novelty-heavy, and the source-candidate embedding component was usually dropped. Held-out major-error recall at a 20\% review budget averaged 54.9\%, with substantial variation across folds. Small learned models, including logistic regression and shallow tree-based models, reached similar but not clearly better performance. Given the limited number of major-error labels, the transparent 0.7 novelty plus 0.3 peer-distance score is preferable to a learned detector at this stage.

\subsection{Backtranslation}

Backtranslation was tested as a separate candidate-aware check. Each English candidate in the 3,000-item calibration set was translated back into Pali, and the backtranslated Pali was compared with the original source. The intuition is that if the English translation omits or distorts source-specific material, a round-trip backtranslation may preserve less of the original Pali wording.

Backtranslation was only performed on the 3,000-item calibration set. This set is not a normal random slice of the corpus. It was deliberately enriched with high-risk items, especially the top-risk tail. That makes the sample easier for many risk scores to rank, because it contains many obvious or semi-obvious high-risk cases. For this reason, the backtranslation results are reported only as within-sample comparisons. They should not be read as full-corpus deployment estimates.

As a standalone metric, chrF backtranslation risk was the strongest backtranslation variant tested. It achieved major-vs-nonmajor AUC 0.824. At a 20\% review budget within the enriched calibration set, it captured 67.1\% of panel-major errors, with 23.2\% major-error precision. This indicates that round-trip lexical loss contains real error signal.

However, backtranslation did not clearly outperform the simpler source-novelty-plus-peer score on the same 3,000 items. In this within-sample comparison, the 0.7 source novelty plus 0.3 peer-centroid score achieved AUC 0.820 and captured 71.0\% of panel-major errors at a 20\% review budget, with 24.5\% major-error precision. Adding chrF backtranslation risk increased AUC only to 0.832 and major-error recall only to 72.5\%, with 25.0\% major-error precision. The gain is therefore small relative to the extra LLM calls required.

Overlap analysis explains the small marginal gain. At the same 20\% budget, chrF backtranslation risk caught 139 major errors, while the source-novelty-plus-peer score caught 147. Of these, 128 were the same errors. Backtranslation therefore appears to be useful supporting evidence, but not a central detector in this experiment. Detailed curves are reported in Appendix Table A3.

\subsection{"Strong" GEMBA Scoring}

No-reference GEMBA scoring, also computed only on the 3,000-item calibration set, was the strongest signal tested. Mean GEMBA risk from the "strong" panel achieved AUC 0.970 for panel-major versus non-major, 0.985 for major versus valid, and 0.910 for any error versus valid. The score distribution separated labels clearly: panel-major errors had mean GEMBA score 69.3, panel-minor errors 87.0, and panel-valid translations 95.2.

At a 5\% review budget, GEMBA mean risk captured 59.4\% of panel-major errors with 82.0\% major-error precision and 98.7\% any-error precision. At 10\%, it captured 81.6\% of panel-major errors with 56.3\% major-error precision. At 20\%, it captured 94.7\% of panel-major errors and 71.9\% of all panel errors. These results are much stronger than the source-prior, embedding-based, and backtranslation results.

\begin{table}[tbp]
\caption{"Strong" GEMBA ranking performance on the 3,000-item calibration set. AUC is a property of the full ranking and therefore repeats across budget cutoffs for the same score; recall and precision are calculated at the listed review budget.}
\begin{center}
\scriptsize
\renewcommand{\arraystretch}{1.15}
\begin{tabularx}{\textwidth}{@{}>{\raggedright\arraybackslash}X >{\raggedright\arraybackslash}X >{\raggedright\arraybackslash}X >{\raggedright\arraybackslash}X >{\raggedright\arraybackslash}X >{\raggedright\arraybackslash}X >{\raggedright\arraybackslash}X >{\raggedright\arraybackslash}X@{}}
\hline
\textit{Metric} & \textit{Major AUC} & \textit{Error-vs-valid AUC} & \textit{Budget} & \textit{Major recall} & \textit{Major precision} & \textit{Any-error recall} & \textit{Any-error precision} \\
\hline
GEMBA mean risk, top 5\% & 0.970 & 0.910 & 5\% & 59.4\% & 82.0\% & 27.2\% & 98.7\% \\
GEMBA mean risk, top 10\% & 0.970 & 0.910 & 10\% & 81.6\% & 56.3\% & 48.2\% & 87.3\% \\
GEMBA mean risk, top 20\% & 0.970 & 0.910 & 20\% & 94.7\% & 32.7\% & 71.9\% & 65.2\% \\
GEMBA minimum-score risk, top 10\% & 0.948 & 0.875 & 10\% & 77.3\% & 53.3\% & 43.8\% & 79.3\% \\
GEMBA score range, top 10\% & 0.810 & 0.774 & 10\% & 45.9\% & 31.7\% & 30.7\% & 55.7\% \\
\hline
\end{tabularx}
\end{center}
\end{table}

A second way to use GEMBA is as a red-flag rule: mark a translation for review if even one "strong" GEMBA scorer gives it a low score. This is different from ranking by the average GEMBA score. With a cutoff of 60 or lower from any scorer, only 4.5\% of the calibration sample was flagged. This small flagged set was very concentrated: 80.1\% of flagged items were panel-major errors and 96.3\% were some kind of panel-labeled error. However, because the rule is strict, it found only 52.7\% of all panel-major errors. A looser cutoff of 70 flagged 6.3\% of the sample and found 65.2\% of panel-major errors, but precision fell to 71.1\%. Thus, a single very low GEMBA score is a strong warning sign, but mean GEMBA risk is better when the goal is to rank all translations under a fixed review budget.

Disagreement alone was less informative. GEMBA score range had AUC 0.810 for major versus non-major, far below mean risk. A range of at least 20 selected 13.9\% of the sample and captured 57.5\% of panel-major errors, but with only 28.5\% major-error precision. High disagreement without an actually low score was especially weak. The main GEMBA signal is therefore low perceived quality, not merely scorer disagreement.

Combining GEMBA with source-novelty-plus-peer routing or backtranslation did not improve over GEMBA mean risk. For example, GEMBA mean risk alone had major AUC 0.970, while GEMBA mean risk plus the source-novelty-plus-peer score had AUC 0.935 and GEMBA mean risk plus backtranslation plus the source-novelty-plus-peer score had AUC 0.920. Once "strong" GEMBA scores are available, the other signals add little to ranking performance. Their main role is upstream: deciding which items should receive higher-effort GEMBA scoring.

\subsection{GEMBA, Source Novelty, and Translator Effects}

One concern was that GEMBA might fail on high-novelty Pali because those same passages are difficult for both translators and judges. The data did not support this concern. Source novelty correlated moderately with GEMBA risk (\texttt{r = 0.408}) and GEMBA score range (\texttt{r = 0.374}), indicating that rarer and less formulaic passages do tend to receive lower and more variable scores. However, GEMBA remained highly discriminative within each novelty band. Mean-risk AUC for panel-major versus non-major was 0.978 in the low-novelty band, 0.949 in the middle band, and 0.960 in the high band.

Adding source novelty to GEMBA worsened performance. Major-error AUC fell from 0.970 for GEMBA mean risk alone to 0.968 with a 0.9/0.1 GEMBA/novelty blend, 0.954 with a 0.7/0.3 blend, and 0.926 with an equal blend. Source novelty also caught few major errors missed by GEMBA at equal budget: at 10\%, source novelty found only seven major errors not already caught by GEMBA; at 20\%, it found only two. Thus, source novelty is useful as a source-side prior, but it should not be used as a heavy correction once GEMBA scores are available. It identifies passages on which machine translation is more likely to fail, not whether a particular translation has failed.

GEMBA also remained strong by translator. "strong" GEMBA mean risk achieved major-vs-nonmajor AUC 0.981 on DeepSeek candidates, 0.968 on Grok candidates, and 0.963 on Qwen candidates. The calibration set contained fewer DeepSeek major errors (31) than Grok (57) or Qwen (119), so per-translator estimates should be read with different uncertainty. Still, the signal is not confined to one candidate model. Full per-translator GEMBA results are reported in Appendix Table A4.

\subsection{Is GEMBA Just Strong-Model Auditing?}

The "strong" GEMBA panel used models that are plausibly stronger than the translator models. To test whether the result was simply a strong-model-judges-weak-models artifact, the same 3,000 calibration items were scored with a "peer" GEMBA panel built from the translator-model families themselves: DeepSeek, Qwen, and Grok, with Kimi 2.6 added so that self-scoring could be excluded while still giving each item three scores. "peer" GEMBA was substantially weaker than "strong" GEMBA, but it still remained clearly useful.

Mean-risk AUC for panel-major versus non-major fell from 0.970 with the "strong" panel to 0.885 with the "peer" panel. Error-versus-valid AUC fell from 0.910 to 0.799. At a 10\% review budget, "strong" GEMBA captured 81.6\% of panel-major errors, while "peer" GEMBA captured 61.4\%. At 20\%, the corresponding values were 94.7\% and 79.2\%.

\begin{table}[tbp]
\caption{"Strong" versus "peer" GEMBA.}
\begin{center}
\scriptsize
\renewcommand{\arraystretch}{1.15}
\begin{tabularx}{\textwidth}{@{}>{\raggedright\arraybackslash}X >{\raggedright\arraybackslash}X >{\raggedright\arraybackslash}X >{\raggedright\arraybackslash}X >{\raggedright\arraybackslash}X >{\raggedright\arraybackslash}X >{\raggedright\arraybackslash}X >{\raggedright\arraybackslash}X@{}}
\hline
\textit{Panel} & \textit{Budget} & \textit{Major AUC} & \textit{Error-vs-valid AUC} & \textit{Major recall} & \textit{Major precision} & \textit{Any-error recall} & \textit{Any-error precision} \\
\hline
"strong" GEMBA & 5\% & 0.970 & 0.910 & 59.4\% & 82.0\% & 27.2\% & 98.7\% \\
"peer" GEMBA & 5\% & 0.885 & 0.799 & 43.5\% & 60.0\% & 23.2\% & 84.0\% \\
"strong" GEMBA & 10\% & 0.970 & 0.910 & 81.6\% & 56.3\% & 48.2\% & 87.3\% \\
"peer" GEMBA & 10\% & 0.885 & 0.799 & 61.4\% & 42.3\% & 36.4\% & 66.0\% \\
"strong" GEMBA & 20\% & 0.970 & 0.910 & 94.7\% & 32.7\% & 71.9\% & 65.2\% \\
"peer" GEMBA & 20\% & 0.885 & 0.799 & 79.2\% & 27.3\% & 56.6\% & 51.3\% \\
\hline
\end{tabularx}
\end{center}
\end{table}

Overlap analysis shows that "peer" GEMBA mostly catches the same major errors as "strong" GEMBA, rather than adding a large independent set. Under the scoring-specific budget rule described above, "strong" GEMBA caught 169 panel-major errors and "peer" GEMBA caught 127 at a 10\% budget; 122 were shared, five were peer-unique, and 47 were strong-unique. At a 20\% budget, "peer" GEMBA added only two major errors not caught by "strong" GEMBA.

Inter-evaluator rank consistency was moderate across both GEMBA panels. The "strong" panel showed its highest rank agreement between GPT-5.5 and Claude Sonnet 4.6 (\texttt{\ensuremath{\rho} = 0.731}), while "peer" panel agreement was weaker and more uneven, especially for Qwen/Grok (\texttt{\ensuremath{\rho} = 0.437}). This supports using averaged GEMBA scores rather than relying on a single scorer, and it reinforces the conclusion that scorer choice matters.

\begin{table}[tbp]
\caption{Pairwise Spearman rank correlations between no-reference GEMBA evaluator scores. Each cell reports Spearman's \texttt{\ensuremath{\rho}}, with the number of co-rated calibration items in parentheses. The "strong" GEMBA panel scored all 3,000 calibration items. The "peer" GEMBA panel excluded self-scoring, so co-rated counts vary. Blank cells indicate evaluator pairs that did not belong to the same GEMBA scoring panel.}
\begin{center}
\scriptsize
\renewcommand{\arraystretch}{1.15}
\begin{tabularx}{\textwidth}{@{}>{\raggedright\arraybackslash}X >{\raggedright\arraybackslash}X >{\raggedright\arraybackslash}X >{\raggedright\arraybackslash}X >{\raggedright\arraybackslash}X >{\raggedright\arraybackslash}X >{\raggedright\arraybackslash}X >{\raggedright\arraybackslash}X@{}}
\hline
\textit{Evaluator} & \textit{GPT-5.5} & \textit{Gemini 3.1 Pro} & \textit{Claude Sonnet 4.6} & \textit{DeepSeek V4 Pro} & \textit{Qwen3.6 Plus} & \textit{Grok 4.3} & \textit{Kimi K2.6} \\
\hline
GPT-5.5 & 1.000\newline{}(3,000) & 0.612\newline{}(3,000) & 0.731\newline{}(3,000) &  &  &  &  \\
Gemini 3.1 Pro & 0.612\newline{}(3,000) & 1.000\newline{}(3,000) & 0.569\newline{}(3,000) &  &  &  &  \\
Claude Sonnet 4.6 & 0.731\newline{}(3,000) & 0.569\newline{}(3,000) & 1.000\newline{}(3,000) &  &  &  &  \\
DeepSeek V4 Pro &  &  &  & 1.000\newline{}(2,013) & 0.595\newline{}(983) & 0.544\newline{}(1,030) & 0.581\newline{}(2,013) \\
Qwen3.6 Plus &  &  &  & 0.595\newline{}(983) & 1.000\newline{}(1,970) & 0.437\newline{}(987) & 0.510\newline{}(1,970) \\
Grok 4.3 &  &  &  & 0.544\newline{}(1,030) & 0.437\newline{}(987) & 1.000\newline{}(2,017) & 0.542\newline{}(2,017) \\
Kimi K2.6 &  &  &  & 0.581\newline{}(2,013) & 0.510\newline{}(1,970) & 0.542\newline{}(2,017) & 1.000\newline{}(3,000) \\
\hline
\end{tabularx}
\end{center}
\end{table}

These results support two claims. First, GEMBA is judge-dependent: stronger scorer models produce a substantially better triage signal. Second, the prompt and scoring method still carry real no-reference information, because the weaker "peer" panel remained meaningfully above the source-prior, embedding-based, and backtranslation signals.

\subsection{Human Author-Anchor Validation}

The 500-item author anchor provides an independent check on the LLM-panel labels and GEMBA risk scores. Because the anchor set is diagnostically enriched, its raw label distribution is not a prevalence estimate. The author labeled 310 items valid, 99 minor errors, and 91 major errors.

The LLM panel showed high sensitivity but conservative overcalling relative to the author labels. Three-class exact agreement was 72.4\%, with Cohen's kappa 0.559. Binary error recall was 97.4\%, meaning that the panel missed very few author-labeled errors. Binary error precision was lower, 64.5\%, because many author-valid items were labeled minor errors by the panel. Major-error recall was 93.4\% and major-error precision 72.6\%.

\begin{table}[tbp]
\caption{Author and LLM-panel labels on the 500-item anchor set. Overall exact author-panel agreement was 362/500 items (72.4\%).}
\begin{center}
\small
\renewcommand{\arraystretch}{1.15}
\begin{tabularx}{\textwidth}{@{}>{\raggedright\arraybackslash}X >{\raggedright\arraybackslash}X >{\raggedright\arraybackslash}X >{\raggedright\arraybackslash}X@{}}
\hline
\textit{Label} & \textit{Author count} & \textit{LLM-panel count} & \textit{Author-panel agreement} \\
\hline
Valid & 310 & 213 & 208 (67.1\%) \\
Minor error & 99 & 170 & 69 (69.7\%) \\
Major error & 91 & 117 & 85 (93.4\%) \\
\hline
\end{tabularx}
\end{center}
\end{table}

Panel-major labels were substantially more reliable than panel-minor labels. Of 117 panel-major items in the anchor set, 85 were author-major, 26 were author-minor, and six were author-valid. In contrast, panel-minor labels included many items the author considered valid. This pattern is important for interpreting the 3,000-item calibration results: the LLM panel is a high-sensitivity triage labeler, not an expert oracle, and it tends to overcall minor errors. At the same time, the boundaries between valid variation and minor error, and between minor and major error, are partly judgment-dependent, especially in interpretively open passages.

Individual judges showed similar performance profiles. GPT-5.5 had exact agreement 73.0\% and kappa 0.569; Gemini 3.1 Pro had exact agreement 71.0\% and kappa 0.525; Claude Sonnet 4.6 had exact agreement 72.6\% and kappa 0.530. GPT-5.5 and Gemini had higher major-error recall, while Claude had somewhat higher binary error precision. Full individual-judge metrics are reported in Appendix Table A5.

\subsection{GEMBA Against the Author Anchor}

"strong" GEMBA remained highly predictive when evaluated against author labels, although less strongly than against LLM-panel labels. Mean "strong" GEMBA risk achieved author-anchor AUC 0.924 for major versus non-major, 0.966 for major versus valid, and 0.931 for error versus valid. "peer" GEMBA was again weaker, with AUC 0.789 for major versus non-major, 0.833 for major versus valid, and 0.772 for error versus valid.

\begin{table}[tbp]
\caption{GEMBA AUC against author labels.}
\begin{center}
\small
\renewcommand{\arraystretch}{1.15}
\begin{tabularx}{\textwidth}{@{}>{\raggedright\arraybackslash}X >{\raggedright\arraybackslash}X >{\raggedright\arraybackslash}X >{\raggedright\arraybackslash}X@{}}
\hline
\textit{Metric} & \textit{Major vs non-major} & \textit{Major vs valid} & \textit{Error vs valid} \\
\hline
"strong" GEMBA mean risk & 0.924 & 0.966 & 0.931 \\
"strong" GEMBA minimum-score risk & 0.904 & 0.951 & 0.907 \\
"strong" GEMBA score range & 0.755 & 0.819 & 0.805 \\
"peer" GEMBA mean risk & 0.789 & 0.833 & 0.772 \\
"peer" GEMBA minimum-score risk & 0.747 & 0.774 & 0.709 \\
"peer" GEMBA score range & 0.674 & 0.702 & 0.649 \\
\hline
\end{tabularx}
\end{center}
\end{table}

Within the author-anchor set, the top 10\% of items by "strong" GEMBA risk captured 47.3\% of author-major errors with 86.0\% major-error precision and 96.0\% any-error precision. Within the same set, the top 20\% captured 73.6\% of author-major errors with 67.0\% major-error precision and 97.0\% any-error precision. These are diagnostic anchor-set results, not deployment-calibrated full-corpus budget estimates. "peer" GEMBA captured fewer author-major errors and had lower precision at the same within-anchor budgets.

\begin{table}[tbp]
\caption{GEMBA mean-risk budget curves against author labels.}
\begin{center}
\scriptsize
\renewcommand{\arraystretch}{1.15}
\begin{tabularx}{\textwidth}{@{}>{\raggedright\arraybackslash}X >{\raggedright\arraybackslash}X >{\raggedright\arraybackslash}X >{\raggedright\arraybackslash}X >{\raggedright\arraybackslash}X >{\raggedright\arraybackslash}X@{}}
\hline
\textit{Panel} & \textit{Budget} & \textit{Major recall} & \textit{Major precision} & \textit{Any-error recall} & \textit{Any-error precision} \\
\hline
"strong" GEMBA & 5\% & 24.2\% & 88.0\% & 13.2\% & 100.0\% \\
"strong" GEMBA & 10\% & 47.3\% & 86.0\% & 25.3\% & 96.0\% \\
"strong" GEMBA & 20\% & 73.6\% & 67.0\% & 51.1\% & 97.0\% \\
"peer" GEMBA & 5\% & 18.7\% & 68.0\% & 13.2\% & 100.0\% \\
"peer" GEMBA & 10\% & 36.3\% & 66.0\% & 25.3\% & 96.0\% \\
"peer" GEMBA & 20\% & 56.0\% & 51.0\% & 44.7\% & 85.0\% \\
\hline
\end{tabularx}
\end{center}
\end{table}

Single-low-score thresholds also remained useful on the author anchor as secondary warning rules. Mean GEMBA remains the main ranker, but a single very low score is operationally easy to interpret. If any "strong" GEMBA scorer assigned a score $\leq$ 60, the selected set contained 111 items (22.2\% of the anchor), captured 75.8\% of author-major errors, and had 62.2\% major-error precision and 90.1\% any-error precision. A stricter threshold of $\leq$ 50 selected 81 items, captured 62.6\% of author-major errors, and had 70.4\% major-error precision. Additional threshold results are reported in Appendix Table A6.

The author-anchor results change the strength but not the direction of the main finding. Against LLM-panel labels, "strong" GEMBA appeared extremely strong. Against author labels, it remains the best reference-free signal tested, but with lower recall at the same within-anchor review fractions. This is a more realistic diagnostic check for human-facing use: "strong" GEMBA is powerful as a triage tool, but its performance depends on scorer strength and it does not eliminate the need for human review.

\subsection{Cross-Method Summary}

Across the tested reference-free signals, the strongest separation came from no-reference GEMBA scoring by the "strong" evaluator panel. Source novelty, peer-centroid distance, and source-novelty-plus-peer composites enriched for major errors but did not approach GEMBA-level precision or recall. Backtranslation added measurable but modest signal over the source-novelty-plus-peer score. "Peer" GEMBA remained useful but was consistently weaker than "strong" GEMBA, showing that evaluator choice substantially affects performance.

The result pattern is therefore ordered rather than binary: source-only and embedding-based signals are useful for triage; backtranslation may be used as supporting evidence; "strong" GEMBA is the best candidate-aware signal tested, while "peer" GEMBA (in case stronger models are not available) is still clearly useful. The discussion sketches how these signals could inform a practical review workflow, while noting that the complete workflow is not evaluated as a single end-to-end cascade.

\section{Discussion}

\subsection{What "No-Reference" Means Here}

The central methodological distinction is that the system is no-reference at inference time, not no-reference in the entire research pipeline. The English translation is used to help produce calibration labels and author-anchor labels, but it is not used to compute any of the risk signals tested for a newly translated passage. This is the condition that makes the results relevant to untranslated or under-translated classical material: the reference functions as a measuring instrument for the experiment, not as an input to the proposed triage system.

This distinction also keeps the study from treating one English translation as the definition of correctness. Reference-based metrics such as BLEU, BERTScore, and COMET have been central to machine translation evaluation \cite{papineni2002bleu,zhang2020bertscore,rei2020comet}, but they presuppose at least one reference translation. For Pali and other classical languages, the motivating use case is precisely where that reference may not exist. Moreover, legitimate translation variation is not noise: translation studies has long emphasized that translation is interpretive, stylistically situated, and shaped by audience and purpose \cite{venuti2008invisibility,tymoczko2007enlarging}.

\subsection{Source Difficulty Is a Risk Prior, Not an Error Detector}

One of the clearest findings is that source novelty is informative. Passages with rarer vocabulary, fewer repeated formulas, and lower neighborhood similarity were more likely to contain panel-major errors. The refined source-novelty-plus-peer score captured an estimated 60.6\% of panel-major errors at a 20\% full-corpus review budget, while source novelty alone captured an estimated 48.7\%. This supports the intuition that some passages are intrinsically riskier for LLM translation before any candidate translation is inspected.

At the same time, source novelty must be interpreted narrowly. It does not detect whether a specific candidate translation is wrong. It estimates that a passage is difficult. This makes it useful for routing: a high-novelty passage may deserve extra scrutiny, multiple translations, or a stronger evaluation model. It does not justify marking the output as erroneous.

This result aligns with recent work on LLM translation of Ancient Greek technical prose. Zainaldin and colleagues found that terminology rarity was a strong predictor of catastrophic translation failure in Galenic texts, especially in passages with dense technical vocabulary \cite{zainaldin2026galen}. Our Pali results are not identical in domain or measurement, but they point in the same direction: in classical-language translation, rare or less formulaic source material can be a substantial risk factor. This is encouraging for generalization, because source rarity and formulaicity are computable in many classical corpora even when no target-language reference exists.

Once "strong" GEMBA scores are available, adding source novelty makes the ranking worse rather than better. This means source novelty belongs upstream as a difficulty prior, not downstream as a correction to a strong candidate-aware evaluator. In practical terms, source novelty helps decide which passages deserve deeper checks; GEMBA helps judge the particular candidate translation.

\subsection{Peer Agreement and Backtranslation Are Useful but Secondary}

Peer-centroid distance is conceptually attractive because it uses other AI translations as a substitute for human references. This resembles multi-hypothesis MT evaluation, where variation among machine outputs can help model translation variability and partially replace reference variation \cite{fomicheva2020multihypothesis}. In our results, peer-centroid distance did add signal: as a single feature it captured an estimated 40.9\% of panel-major errors at a 20\% full-corpus review budget. Combined with source novelty, it contributed to the best transparent source-novelty-plus-peer routing score.

However, peer-centroid distance is not a cheap signal in the ordinary sense. It requires multiple translations of the same passage, plus embeddings. It is reference-free, but not cost-free. This matters for deployment. If a translation workflow already produces multiple model outputs, peer distance is a natural by-product. If it produces only one candidate, peer distance becomes an additional translation expense.

Backtranslation also carried signal, but only modestly. Round-trip lexical loss is intuitively relevant: if an English translation drops source-specific Pali material, a backtranslation may fail to recover it. In this study, chrF backtranslation risk plus the source-novelty-plus-peer score improved major-vs-nonmajor AUC from 0.820 to 0.832. At a 20\% calibration-sample review budget, major recall increased only from 71.0\% to 72.5\%. This is evidence that backtranslation sees something real, but not enough to make it a central detector.

This finding is useful because it prevents overbuilding the system. Backtranslation is expensive, and its marginal gain was small. It may still be valuable in targeted settings, especially when a human reviewer wants another view of what semantic material survived the candidate translation. But as a scoring layer, it should remain supporting evidence rather than a coequal partner to GEMBA.

\subsection{GEMBA Works, but It Is Judge-Dependent}

The strongest candidate-aware signal was no-reference GEMBA. This is consistent with Kocmi and Federmann's finding that prompted LLMs can be strong evaluators of translation quality in both reference-based and reference-free modes \cite{kocmi2023gemba}. In our calibration set, "strong" GEMBA mean risk reached AUC 0.970 for panel-major versus non-major and captured 81.6\% of panel-major errors at a 10\% review budget. These numbers are much stronger than source novelty, peer distance, or backtranslation.

The author anchor makes this result more credible but also more realistic. Against author labels, "strong" GEMBA mean risk remained the best signal tested, with AUC 0.924 for major versus non-major and 0.966 for major versus valid. However, within the author-anchor set, the top 10\% by "strong" GEMBA risk captured only 47.3\% of author-major errors. This gap is expected: the 3,000-item calibration labels are produced by an LLM panel, while the anchor checks those labels and scores against human judgment. The anchor therefore lowers the apparent strength of GEMBA, but it does not reverse the conclusion.

The "strong" versus "peer" GEMBA comparison also matters. When the scorer panel was changed from GPT-5.5, Gemini 3.1 Pro, and Claude Sonnet 4.6 to a "peer" panel closer to the translator set, performance dropped substantially. Major AUC fell from 0.970 to 0.885 against panel labels, and from 0.924 to 0.789 against author labels. This confirms that GEMBA is not model-independent. The prompt matters, but scorer strength matters too.

The appropriate conclusion is therefore not that "GEMBA is gold." It is narrower and more targeted: "strong" GEMBA is the best candidate-aware no-reference triage signal tested, but its performance depends on scorer strength and it does not remove the need for human review. This conclusion also fits the broader LLM-as-judge literature. Strong LLM judges can approximate human preferences surprisingly well in some settings, but documented biases and instability require calibration and human anchoring \cite{zheng2023judging,wang2024fair}.

\subsection{The LLM Panel Is a Triage Labeler, Not an Oracle}

The author-anchor analysis is important because it prevents circularity. If the only labels were LLM-panel labels, "strong" GEMBA could be criticized as merely predicting judgments made by related LLMs. The anchor shows that the panel and GEMBA scores are meaningfully aligned with author judgment, but also that the panel overcalls errors, especially minor ones.

This should be framed as a feature of the study design rather than a failure. The panel is used as a high-sensitivity triage labeler. It missed very few author-labeled errors in the anchor set: binary error recall was 97.4\%, and major-error recall was 93.4\%. The cost was lower precision. Many author-valid translations were labeled minor errors by the panel, and panel-minor labels were much less reliable than panel-major labels.

For a quality-oriented translation audit, this asymmetry is acceptable and even desirable. Missing a major error in a religious or philosophical text is more serious than over-referring a defensible variation for review. But it also means that panel-derived prevalence estimates should be interpreted as triage-prevalence estimates, not as final expert error rates.

\subsection{Limitations}

The study has several limitations. First, the system is no-reference at inference time, but not no-reference in its research design. Sujato's English translation is used to support LLM and author adjudication. This is appropriate for calibration, but it means that the reported performance depends partly on the suitability of that reference-aided adjudication process.

Second, the 3,000-item calibration labels are produced by an LLM panel. The 500-item author anchor reduces the risk of circularity, but it is still a single-author check rather than an independent expert consensus panel. The author has formal Pali training, but is not an independent external expert, and the study does not measure inter-rater reliability among multiple Pali-trained readers. For that reason, panel-derived prevalence estimates should be read as triage-prevalence estimates, not final expert error rates.

Third, the best-performing signal, "strong" GEMBA, is model-dependent and computationally expensive. Its performance fell when "strong" scorer models were replaced with "peer" scorer models closer to the translator set. The conclusion is therefore not that any LLM can reliably judge any classical-language translation, but that "strong" no-reference LLM scoring can be highly informative when calibrated and human-anchored. In real-world deployments, however, the strongest available models may already be used for translation, leaving no clearly stronger evaluator for GEMBA scoring. In that setting, "peer" GEMBA may still be useful, but its weaker performance should be expected.

Fourth, the experiment is limited to Pali-to-English, three candidate translator models, one large segmented corpus, and passage-level translation. Some Pali passages depend on broader discourse context, formulaic ellipsis, or traditional interpretive background that a passage-level prompt may not provide. The findings should therefore be tested on other classical languages and on other text types before being treated as a general property of LLM translation.

Finally, AI translation of religious and philosophical texts has ethical stakes beyond technical accuracy. The proposed system is a tool for allocating review attention, not for assigning interpretive authority. Deployment on living traditions should include human translators, scholars, and relevant communities in deciding what counts as acceptable translation, what errors matter most, and how AI-generated text should be presented to readers. Any AI-assisted workflow in religious or philosophical domains should serve accessibility by broadening access to texts for which no human translation exists, not by replacing human expertise.

\subsection{Implications for Classical-Language Translation Workflows}

The results suggest a practical, but not yet end-to-end validated, workflow for AI-assisted classical-language translation. First, compute source novelty over the source corpus to identify passages that are likely to be difficult. Second, if the workflow permits multiple candidate translations, compute peer-centroid distance to identify outputs that diverge from other renderings. Third, use "strong" no-reference GEMBA scoring for the subset where review resources are available. Fourth, route high-risk items to human review, retranslation, or both.

This workflow is not a replacement for philological expertise. It is a workload allocation system. Its value is highest when the corpus is large, expert time is scarce, and the cost of missing a major error is high. That description fits many classical-language settings: Pali, Latin, Ancient Greek, Sanskrit, Coptic and other traditions where large bodies of untranslated texts exist but expert translators are few.

Current LLM translation work on classical languages is moving quickly. LITERA, for example, frames Latin-to-English LLM translation as a research-assistance system rather than a fully autonomous replacement for translators \cite{rosu2025litera}. The Ancient Greek Galen study similarly shows that LLMs can produce high-quality translations for some passages while failing sharply on rare technical material \cite{zainaldin2026galen}. Our Pali results fit this emerging pattern. LLMs are useful enough to justify serious evaluation infrastructure, but not reliable enough to be left unaudited.

The most general contribution of the present study is therefore methodological. The proposed design does not require human references for the texts being newly translated. It requires source text, candidate translations, and a calibration set where labels can be obtained with reference aid or human expertise. Once calibrated, the same pattern can be transferred: source-side risk prior, candidate-aware no-reference signals, "strong" GEMBA scoring, and human review of the flagged tail. For classical-to-modern translation, that may be a more realistic goal than fully automatic quality certification.

A natural next step is to turn the calibrated labels into a dedicated quality-estimation model for Pali or related classical languages. Domain-specific low-resource QE research suggests that lightweight adaptation can improve robustness when prompt-only evaluation is insufficient \cite{gurav2026quality}. The present study does not train such a model; it provides the kind of labeled triage data and error-risk analysis that would make that future work possible.
\clearpage

\section*{Acknowledgments}
The author gratefully acknowledges SuttaCentral and Bhikkhu Sujato for their translation work and for making Pali texts and English translations accessible in a structured digital format. Their work made this study possible.

\section*{Funding}

This study received no external funding and was self-funded by the author.

\section*{Data Availability}

The Pali source texts and Bhikkhu Sujato's corresponding English reference translations are available through SuttaCentral's public GitHub repository at \url{https://github.com/suttacentral/bilara-data}; Sujato's translations are released under a Creative Commons Zero (CC0) licence. A reproducibility package for this study, including the full underlying data generated for the analysis, LLM translations, scores, and scripts, is available at \url{https://github.com/MateMetzger/pali-translation-error-triage}.

\section*{Generative AI Use Statement}

Generative AI was used in two distinct ways in this study. First, LLM outputs are the object of study: the research evaluates and triages AI-generated Pali-to-English translations. Second, generative AI, specifically GPT-5.5, was used as an assistive aid for phrasing, language editing, polishing, and code generation during the research and manuscript preparation process. Generative AI was not used to autonomously generate manuscript sections or to independently interpret the research data or derive conclusions. All AI-assisted outputs were reviewed by the author, who takes responsibility for the content of the manuscript.

\bibliographystyle{plainnat}
\bibliography{bibliography}

\clearpage

\appendix
\section{Supplementary Tables}

\setcounter{table}{0}
\renewcommand{\thetable}{A\arabic{table}}
\renewcommand{\theHtable}{A\arabic{table}}

\begin{table}[!htbp]
\caption{Final corpus by source novelty band.}
\begin{center}
\scriptsize
\renewcommand{\arraystretch}{1.15}
\begin{tabularx}{\textwidth}{@{}>{\raggedright\arraybackslash}X >{\raggedright\arraybackslash}X >{\raggedright\arraybackslash}X >{\raggedright\arraybackslash}X >{\raggedright\arraybackslash}X >{\raggedright\arraybackslash}X@{}}
\hline
\textit{Source novelty band} & \textit{Passages} & \textit{Segments} & \textit{Pali characters} & \textit{Sujato characters} & \textit{Novelty range} \\
\hline
Low & 5,526 & 34,838 & 2,812,173 & 2,846,041 & -7.812 to -0.385 \\
Middle & 5,697 & 32,084 & 2,363,188 & 2,286,894 & -0.385 to 0.229 \\
High & 4,270 & 24,552 & 1,302,811 & 1,360,729 & 0.230 to 3.295 \\
\hline
\end{tabularx}
\end{center}
\end{table}

\begin{table}[!htbp]
\caption{Full-corpus review estimates for broad initial-risk boundaries.}
\begin{center}
\scriptsize
\renewcommand{\arraystretch}{1.15}
\begin{tabularx}{\textwidth}{@{}>{\raggedright\arraybackslash}X >{\raggedright\arraybackslash}X >{\raggedright\arraybackslash}X >{\raggedright\arraybackslash}X >{\raggedright\arraybackslash}X >{\raggedright\arraybackslash}X@{}}
\hline
\textit{Reviewed region} & \textit{Reviewed instances} & \textit{Major recall} & \textit{Major precision} & \textit{Any-error recall} & \textit{Any-error precision} \\
\hline
Top 20\% & 9,296 & 40.0\% & 5.9\% & 30.5\% & 18.6\% \\
Top 40\% & 18,592 & 63.2\% & 4.6\% & 50.8\% & 15.5\% \\
Top 60\% & 27,888 & 72.7\% & 3.6\% & 68.2\% & 13.9\% \\
Top 80\% & 37,184 & 87.7\% & 3.2\% & 83.6\% & 12.7\% \\
\hline
\end{tabularx}
\end{center}
\end{table}

\begin{table}[!htbp]
\caption{Backtranslation metrics compared with the refined source-novelty-plus-peer score.}
\begin{center}
\scriptsize
\renewcommand{\arraystretch}{1.15}
\begin{tabularx}{\textwidth}{@{}>{\raggedright\arraybackslash}X >{\raggedright\arraybackslash}X >{\raggedright\arraybackslash}X >{\raggedright\arraybackslash}X >{\raggedright\arraybackslash}X >{\raggedright\arraybackslash}X >{\raggedright\arraybackslash}X >{\raggedright\arraybackslash}X@{}}
\hline
\textit{Metric} & \textit{Major AUC} & \textit{Error-vs-valid AUC} & \textit{Budget} & \textit{Major recall} & \textit{Major precision} & \textit{Any-error recall} & \textit{Any-error precision} \\
\hline
Refined source-novelty-plus-peer score & 0.820 & 0.739 & 10\% & 47.3\% & 32.7\% & 30.5\% & 55.3\% \\
chrF backtranslation risk & 0.824 & 0.745 & 10\% & 46.4\% & 32.0\% & 30.1\% & 54.7\% \\
chrF backtranslation + source-novelty-plus-peer score & 0.832 & 0.753 & 10\% & 50.7\% & 35.0\% & 31.6\% & 57.3\% \\
Refined source-novelty-plus-peer score & 0.820 & 0.739 & 20\% & 71.0\% & 24.5\% & 51.7\% & 46.8\% \\
chrF backtranslation risk & 0.824 & 0.745 & 20\% & 67.1\% & 23.2\% & 46.7\% & 42.3\% \\
chrF backtranslation + source-novelty-plus-peer score & 0.832 & 0.753 & 20\% & 72.5\% & 25.0\% & 52.2\% & 47.3\% \\
\hline
\end{tabularx}
\end{center}
\end{table}

\begin{table}[!htbp]
\caption{"strong" GEMBA by candidate translator.}
\begin{center}
\scriptsize
\renewcommand{\arraystretch}{1.15}
\begin{tabularx}{\textwidth}{@{}>{\raggedright\arraybackslash}X >{\raggedright\arraybackslash}X >{\raggedright\arraybackslash}X >{\raggedright\arraybackslash}X >{\raggedright\arraybackslash}X >{\raggedright\arraybackslash}X@{}}
\hline
\textit{Candidate translator} & \textit{Calibration n} & \textit{Panel-major n} & \textit{Major AUC} & \textit{Major-vs-valid AUC} & \textit{Error-vs-valid AUC} \\
\hline
DeepSeek & 987 & 31 & 0.981 & 0.988 & 0.925 \\
Grok & 983 & 57 & 0.968 & 0.985 & 0.893 \\
Qwen & 1,030 & 119 & 0.963 & 0.984 & 0.913 \\
\hline
\end{tabularx}
\end{center}
\end{table}

\begin{table}[!htbp]
\caption{Individual LLM judges versus author labels.}
\begin{center}
\scriptsize
\renewcommand{\arraystretch}{1.15}
\begin{tabularx}{\textwidth}{@{}>{\raggedright\arraybackslash}X >{\raggedright\arraybackslash}X >{\raggedright\arraybackslash}X >{\raggedright\arraybackslash}X >{\raggedright\arraybackslash}X >{\raggedright\arraybackslash}X >{\raggedright\arraybackslash}X@{}}
\hline
\textit{Judge} & \textit{Exact agreement} & \textit{Kappa} & \textit{Error recall} & \textit{Error precision} & \textit{Major recall} & \textit{Major precision} \\
\hline
GPT-5.5 & 73.0\% & 0.569 & 98.9\% & 65.3\% & 94.5\% & 69.9\% \\
Gemini 3.1 Pro & 71.0\% & 0.525 & 93.2\% & 66.0\% & 94.5\% & 65.2\% \\
Claude Sonnet 4.6 & 72.6\% & 0.530 & 87.4\% & 70.3\% & 81.3\% & 65.5\% \\
\hline
\end{tabularx}
\end{center}
\end{table}

\begin{table}[!htbp]
\caption{Single-low-score warning thresholds against author labels.}
\begin{center}
\scriptsize
\renewcommand{\arraystretch}{1.15}
\begin{tabularx}{\textwidth}{@{}>{\raggedright\arraybackslash}X >{\raggedright\arraybackslash}X >{\raggedright\arraybackslash}X >{\raggedright\arraybackslash}X >{\raggedright\arraybackslash}X >{\raggedright\arraybackslash}X >{\raggedright\arraybackslash}X@{}}
\hline
\textit{Panel} & \textit{Rule} & \textit{Selected} & \textit{Major recall} & \textit{Major precision} & \textit{Any-error recall} & \textit{Any-error precision} \\
\hline
"strong" & Any score $\leq$ 50 & 81 (16.2\%) & 62.6\% & 70.4\% & 38.9\% & 91.4\% \\
"strong" & Any score $\leq$ 60 & 111 (22.2\%) & 75.8\% & 62.2\% & 52.6\% & 90.1\% \\
"strong" & Any score $\leq$ 70 & 141 (28.2\%) & 80.2\% & 51.8\% & 65.3\% & 87.9\% \\
"strong" & Any score $\leq$ 80 & 192 (38.4\%) & 91.2\% & 43.2\% & 78.9\% & 78.1\% \\
"peer" & Any score $\leq$ 50 & 140 (28.0\%) & 58.2\% & 37.9\% & 48.4\% & 65.7\% \\
"peer" & Any score $\leq$ 60 & 160 (32.0\%) & 64.8\% & 36.9\% & 53.2\% & 63.1\% \\
"peer" & Any score $\leq$ 70 & 211 (42.2\%) & 78.0\% & 33.6\% & 63.2\% & 56.9\% \\
\hline
\end{tabularx}
\end{center}
\end{table}
\clearpage

\end{document}